\documentclass{article}

\usepackage{times}
\usepackage{calc}
\usepackage{float}
\usepackage[numbers,sort]{natbib}
\usepackage{PRIMEarxiv}
\usepackage[super]{nth}
\usepackage[utf8]{inputenc} 
\usepackage[T1]{fontenc}    
\usepackage{url}            
\usepackage{booktabs}       
\usepackage{amsfonts}       
\usepackage{nicefrac}       
\usepackage{microtype}      
\usepackage{lipsum}
\usepackage{fancyhdr}       
\usepackage{graphicx}       
\usepackage{atbegshi}
\graphicspath{{media/}}     
\title{An Intensity and Phase Stacked Analysis of Phase-OTDR System using Deep Transfer Learning and Recurrent Neural Networks}

\newcommand\blfootnote[1]{%
  \begingroup
  \renewcommand\thefootnote{}\footnote{#1}%
  \addtocounter{footnote}{-1}%
  \endgroup
}
\author{\large
  Ceyhun Efe Kayan, Kivilcim Yuksel Aldogan, Abdurrahman Gumus*
  \\\large
  Department of Electrical and Electronics Engineering, İzmir Institute of Technology}

\begin{document}
\maketitle

\begin{abstract}
Distributed acoustic sensors (DAS) are effective apparatus which are widely used in many application areas for recording signals of various events with very high spatial resolution along the optical fiber. To detect and recognize the recorded events properly, advanced signal processing algorithms with high computational demands are crucial. Convolutional neural networks are highly capable tools for extracting spatial information and very suitable for event recognition applications in DAS. Long-short term memory (LSTM) is an effective instrument for processing sequential data. In this study, we proposed a two stage feature extraction methodology that combines the capabilities of these neural network architectures with transfer learning to classify vibrations applied to an optical fiber by a piezo transducer. First, we extracted the differential amplitude and phase information from the $\Phi$-OTDR recordings and stored them in a temporal-spatial data matrix. Then, we used a state-of-the-art pre-trained CNN without dense layers as a feature extractor in the first stage. In the second stage, we used LSTMs to further analyze the features extracted by the CNN. Finally, we used a dense layer to classify the extracted features. To observe the effect of the utilized CNN architecture, we tested our model with five state-of-the art pre-trained models (VGG-16, ResNet-50, DenseNet-121, MobileNet and Inception-v3). The results show that using the VGG-16 architecture in our framework manages to obtain 100\% classification accuracy in 50 trainings and got the best results on our $\Phi$-OTDR dataset. Outcomes of this study indicate that the pre-trained CNNs combined with LSTM are very suitable for the analysis of differential amplitude and phase information, represented in a temporal spatial data matrix which is promising for event recognition operations in DAS applications.\blfootnote{* Corresponding author: abdurrahmangumus@iyte.edu.tr \\ \\  Preprint submitted to Springer. Under review.}
\end{abstract}


\section{Introduction}
Fiber-based distributed acoustic sensors (DAS) are powerful instruments that have been widely used in successful application areas across many sectors \cite{Gorshkov2022}. These areas range from the most popular applications of perimeter security (e.g., railways, airports, power plants, pipelines, military fields, etc.) \cite{Juarez2005, Aktas2017, Ozkan2020} to geophysical studies (e.g., vertical seismic profiling \cite{Mateeva2014}, monitoring the deep ocean \cite{Hartog2019}, subsurface imaging in boreholes \cite{Bakulin2017}). DAS have been rapidly evolving to include even more purely scientific or exotic applications (biology \cite{s21051592}, natural sciences \cite{Caruso2020}, humanitarian sciences, and culture \cite{Golacki2016}). This interest in fiber-based systems is largely due to the vast potential of DAS to become an essential sensing technology by converting the existing communication fibers into unprecedented large-scale pervasive sensor networks \cite{Wang2020}.

Phase-sensitive optical time-domain reflectometer ($\Phi$-OTDR) is a common implementation of fiber-based DAS, in which optical pulses emitted from a highly coherent laser source are launched into the sensing fiber to perform spatially-resolved measurements of Rayleigh backscattering signal (RBS). $\Phi$-OTDR schemes can be divided into two main categories according to the detection method employed: direct detection and coherent detection \cite{9203322}. In the former case, vibration location and frequency are obtained from the RBS power variation over time, whereas the latter adopts the coherent detection scheme to extract the phase component (proportional to the magnitude of external vibrations) of the RBS, allowing a quantitative demodulation capability.

The enormous potential of DAS implementations in the marketplace resulted in the steadily expanding commercialization of $\Phi$-OTDR interrogators in the last decades. Still, inherently weak RBS signal (low SNR) and fading points (or sensitivity decrease) at random positions all along the fiber are the main problems for especially long-range systems. Sensitivity improvement for DAS has been investigated in two manners. The first one consists of various -hardware or signal processing- enhancements on the interrogator side, whereas the second approach is based on realizing some -distributed or quasi-distributed- modifications on the sensing fiber itself (the readers may refer to \cite{9203322} for more explanations on these two approaches).

In addition to fading noise which impacts the detection performance, there are some other technical challenges of DAS such as environmental interferences (temperature, strains), low-frequency noise (noisy ambient), the absence of multi-dimensional spatial position information, and external disturbances (digging, walking and vehicle passing if the application does not aim to measure them). Due to these major performance limitations, event detection and classification tasks are still considered as challenging tasks for $\Phi$-OTDR systems. Additionally, the requirement of signal processing steps and the amount of data are the bottlenecks for the data analysis, especially for real-time applications. In recent years, several different machine learning models are applied to $\Phi$-OTDR data for handling particular tasks such as denoising, event detection, and event classification.

In the literature, various methods are used to extract features from $\Phi$-OTDR data for recognizing different events, such as wavelet decomposition \cite{Jiang2010}, Gaussian mixture model \cite{Min2007}, and Mel-frequency coefficients \cite{Zhang2015}. By using these signal-related feature extraction methods, recognition tasks were achieved with sufficient performance. However, the requirement of preprocessing steps for feature extraction was still a limiting factor. Moreover, the coherent-fading phenomenon is not totally overcome as it requires additional work to avoid certain points where the sensitivity is severely degraded. Genuinely, all of the mentioned methods use temporal information for handling recognition tasks. Notwithstanding, the data collected with $\Phi$-OTDR also contains information in the spatial domain. In order to process the spatial information, Sun $\emph{et al.}$ proposed an event classification method that uses relevance vector machines to classify morphologic features which are extracted from a temporal-spatial data matrix \cite{Sun2015}. This study indicates that temporal-spatial data matrices are more capable of extracting information in event classification tasks. On the other hand, the extraction of the morphologic features is a complex issue that has high computational demands making it challenging for real-time applications.

$\Phi$-OTDR is sensitive to various parameters related to the environment and the sensing setup. Therefore, using signal-feature-based methods is insufficient in many field applications as it is difficult to find common features. On the other hand, convolutional neural network (CNN) based solutions are robust to these environment-related problems since they are capable of adapting their parameters accordingly. Some researchers have used CNN-based models for handling classification tasks \cite{Shi2019, Wu2021,Pan2010}. Their studies showed that CNNs are eligible tools for extracting features from a temporal-spatial data matrix since it is a two-dimensional representation of the signal. Shi $\emph{et al.}$ converted temporal-spatial data matrices into grayscale images and managed to classify five kinds of events by using a CNN \cite{Shi2019}. In this work, a well-known CNN architecture was optimized, where bandpass filtering and greyscale image transformation were the only preprocessing steps implemented. This particular work shows that the CNNs are proper for extracting the information in the spatial domain. Another methodology applied for handling event classification tasks is the usage of phase information in the signal. Wu $\emph{et al.}$ proposed a multi-input single-output model to ultimately use the collected information for enhancing the classification accuracy \cite{Wu2021}. In the study, they used the Mel-spectrograms of the intensity and phase waveforms as inputs of two different CNNs and they concatenated the extracted features for the classification of the events. In this way, the usage of both intensity and phase was shown to improve the classification accuracy.

Despite many advantages provided by CNNs, their training requires a large amount of data samples for classifying events with good accuracy along with a demand for high computational power. Transfer learning is a famous methodology in the computer vision domain to overcome these problems \cite{Pan2010}. It can be defined as the usage of pre-trained deep CNN architectures for extracting features. Since pre-trained CNNs are trained with an enormous amount of data by deep learning specialists, they are capable of avoiding overfitting. In the literature, many of the researchers utilized feature extractors that are trained in a different domain for obtaining sufficient information from data. Their studies express that the pre-trained feature extractors provide significant performances with remarkable benefits. As an example, Narin $\emph{et al.}$ used pre-trained CNNs for extracting features from chest X-Ray images and trained a feedforward neural network for classifying COVID-19 cases \cite{Narin2021}. Pre-trained CNNs were also used for extracting features from scalograms (combined with SVM classifiers) for schizophrenia detection \cite{Shalbaf2020} and from EEG images for the detection of neonatal seizures \cite{Caliskan2021}. Recently, the pre-trained AlexNet architecture with an SVM classifier was used to classify different events recorded by a $\Phi$-OTDR setup \cite{Li2022}.

In this study, we proposed a transfer learning-based new framework to classify disturbances on a fiber optic cable by using pre-trained deep CNNs. The disturbances of various frequencies and amplitudes were applied by a PZT (piezo transducer) on a 40-meter section of the sensing fiber. A coherent detection $\Phi$-OTDR was used for the interrogation. When the particular fiber section under the influence of PZT is dynamically strained, the refractive index and relative positions of the Rayleigh scattering centers change. As a result, the amplitude and phase of the photo-detected beat signal vary in response to the refractive index change along the fiber. The amplitude and phase evalution of the $\Phi$-OTDR traces were calculated by series of algorithms explained in the Preprocessing section. Then, both phase and differential amplitude data were represented in a temporal-spatial data matrix form. Matrices created in this way were converted into RGB images. 

In the proposed framework, constructed images are fed into five pre-trained deep CNNs whose fully connected layers are removed. Deep CNNs, which are developed and trained by machine learning specialists for the ImageNet Large Scale Visual Recognition Challenge (ILSVRC), are used. Outputs of the CNNs are reshaped to be one-dimensional vectors and each vector is passed to the batch normalization layer. After normalization, feature vectors of two images are concatenated. In order to extract the information in the time domain, concatenated features are fed into bidirectional LSTM layers which are capable of analyzing sequential data. In the end, the output of the bidirectional LSTM layers is directed to a softmax layer to predict the class of the disturbance. Since transfer learning is employed, only the bidirectional LSTM and Softmax layers are trained. In order to observe the performance of different pre-trained deep CNNs, we used five different CNN architectures which are well-known in the computer vision domain including VGG16, Res-Net50, DenseNet121, MobileNet, and Inception-v3. While developing our framework, we utilized TensorFlow’s Keras backend.

The key novelties of our study can be summarized as follows:
\begin{itemize}
    \item[\textbf{(i)}]  Both differential amplitude and phase information are represented in a temporal-spatial data matrix form and created matrices are converted into RGB images.
    \item[\textbf{(ii)}]  Feature extraction from differential amplitude and phase images is achieved by using pre-trained deep CNNs. With the help of transfer learning, overfitting, and other difficulties of training CNNs are avoided.
    \item[\textbf{(iii)}]  In order to process the time-related information, bidirectional LSTM layers are used on the features extracted by pre-trained deep CNNs.
    \item[\textbf{(iv)}]  By employing transfer learning, we developed a system with low computational demands. The proposed framework is trained by using a GTX-level GPU with an exceptionally low training time. 
    \item[\textbf{(v)}]  We worked with a limited and small amount of data samples and achieved to get high classification accuracy.
\end{itemize}

The rest of the paper is organized as follows. In the first section, conducted measurements using a $\Phi$-OTDR setup, principles of the applied methodology, and details of the implemented deep learning model are reported. Then, the experimental results and related discussions are presented. The conclusion of the study is reported in the last section.

\section{Methods}
\label{sec:methods}

\subsection{Dataset}
\subsubsection{Phase-OTDR Experiments}
\label{sec:experiment}
\begin{figure}[h]
    \centering
    \includegraphics[scale=1.5]{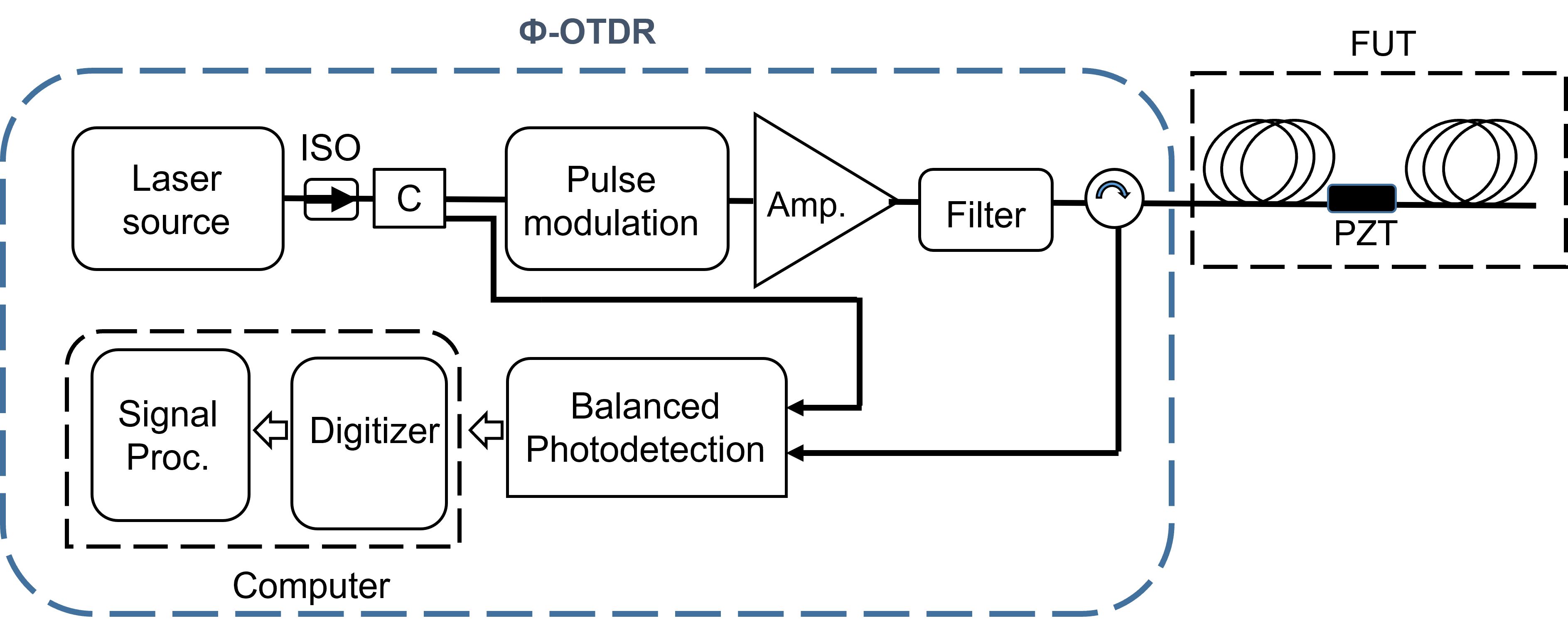}
    \caption{Scheme of the coherent $\Phi$-OTDR set-up. ISO: Optical Isolator, FUT: Fiber Under Test, PZT: Piezo-electric Transducer}
    \label{fig:experiment}
\end{figure}

The coherent detection-based $\Phi$-OTDR set-up implemented in this work is represented in Figure \ref{fig:experiment}. It consists of a highly coherent light source (having 0.1 kHz linewidth, 40 mW continuous output power, and a center wavelength of 1552.51 nm) followed by a pulse modulation stage that outputs narrow optical pulses (of 100 ns pulse width). After amplification and filtering (elimination of noise from the optical amplifier), pulses are launched into the sensing fiber, where the PZT is placed after a lead fiber of 2.3 km. Prior to the data acquisition, an optical demodulation scheme handles the signal reflected from the sensing fiber to extract the phase information. Here, a balanced photodetection scheme is employed, where the Rayleigh backscattered light wave is combined with a small portion (10\%) of the CW laser signal (acting as a reference signal, or local oscillator, LO) using a 50:50 coupler and the AC component of the beat signal is detected by a balanced photodetector. The electrical signal at the output of the balanced photodetection module is sampled by a 1 GHz digitizer (samples are separated by 0.1 m of sampling resolution). Each $\Phi$-OTDR trace of N samples (N=50000) is taken between successive pulses (pulse repetition frequency PRF is about 20 kHz). Many consecutive traces were recorded for each measurement resulting in a slow-time window of about 30 ms (the total measurement time, given as the number of pulses times the pulse separation).

The total length of Fiber Under Test (FUT) is about 4.7 km. The perturbations were applied by a PZT acting on 40 m of wound single-mode fiber located between 2320 and 2360 m along with the sensing fiber. The signal applied to drive the PZT is a sinusoidal wave whose peak-to-peak voltage (${A_{PZT}}$) can be varied between 0 and 10 V and frequency (${f_{vib}}$) adjusted from 200 Hz to 9 kHz. A detailed discussion of the experimental procedure and the complete list of parameters used in the experiments are given in our previous work \cite{Aldogan2018}. In order to verify the data analysis method proposed in this work, a series of $\Phi$-OTDR traces (vibration events) have been used, having three different vibration amplitudes (${A_{pzt} = 1, 2, 3}$V), and five different vibration frequencies (${f_{vib}= 50, 100, 200, 500, 1000}$ Hz).

\subsubsection{Preprocessing}
\label{sec:preprocessing}
In order to create a temporal-spatial data matrix, consecutive $\Phi$-OTDR traces are separated from the recording array and placed as rows in the matrix. In this way, rows in the matrix represent the spatial distribution and each column presents the variation of the trace at a particular position during the slow-time window. Then the following procedures are followed to obtain the final temporal-spatial matrices:
\begin{itemize}
    \item A bandpass filtering is applied for collecting the frequency components centered around the pulse modulator’s frequency shift (an acousto-optic modulator was used in the set up imposing a frequency shift of ${F_{AOM}= 160}$ MHz \cite{Aldogan2018}).
    \item Each term in a $\Phi$-OTDR trace is converted into a complex number by applying the Hilbert transform.
    \item By using the magnitude operation, the amplitude matrix is obtained.
    \item To obtain the phase matrix, the phase shift of the acousto-optic modulator is subtracted from the angle of each complex term.
    \item The differential amplitude matrix is obtained by subtracting the first row of the amplitude matrix from the other columns.
    \item The first column of the phase matrix is subtracted from other columns to remove the “initial phase” caused by the laser source.
    \item The mode of each term in the phase matrix with 2$\pi$ is calculated.
\end{itemize}
Since the phase values were increasing with the distance linearly, the information related to the phase was disappearing. Due to this incremental behavior, phase unwrapping operation was applied. As an example, $\pi$ and 3$\pi$ are both correspond to a phase of $\pi$. So, in order to fix this issue, the mode of each term with 2$\pi$ is calculated, and the phase values are squeezed into the [0, 2$\pi$] interval.
After these operations, each matrix is saved as an image with a resolution of 1000 $\times$ 1000.

\begin{figure}[h]
    \centering
    \includegraphics[scale=1.5]{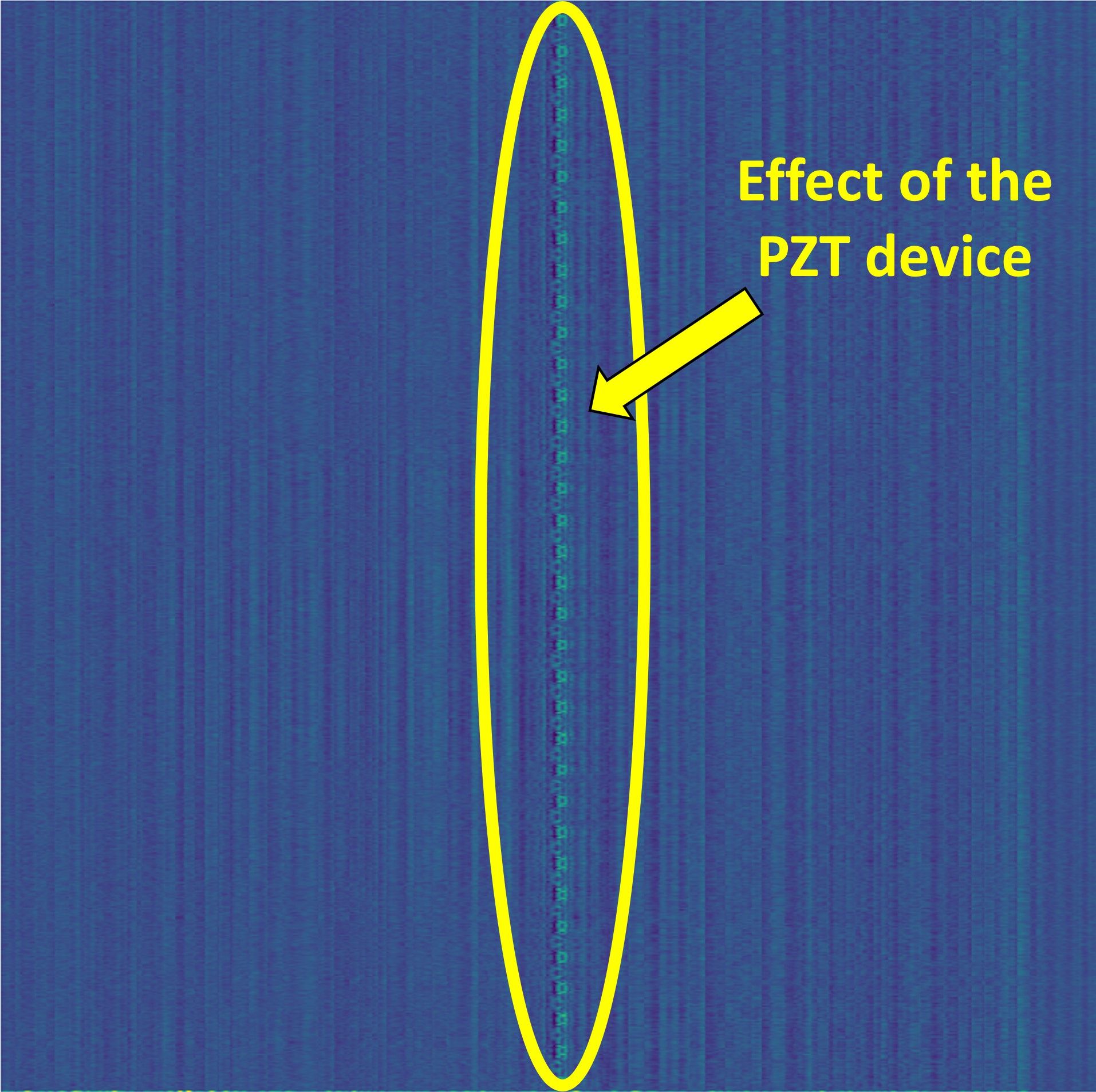}
    \includegraphics[scale=1.5]{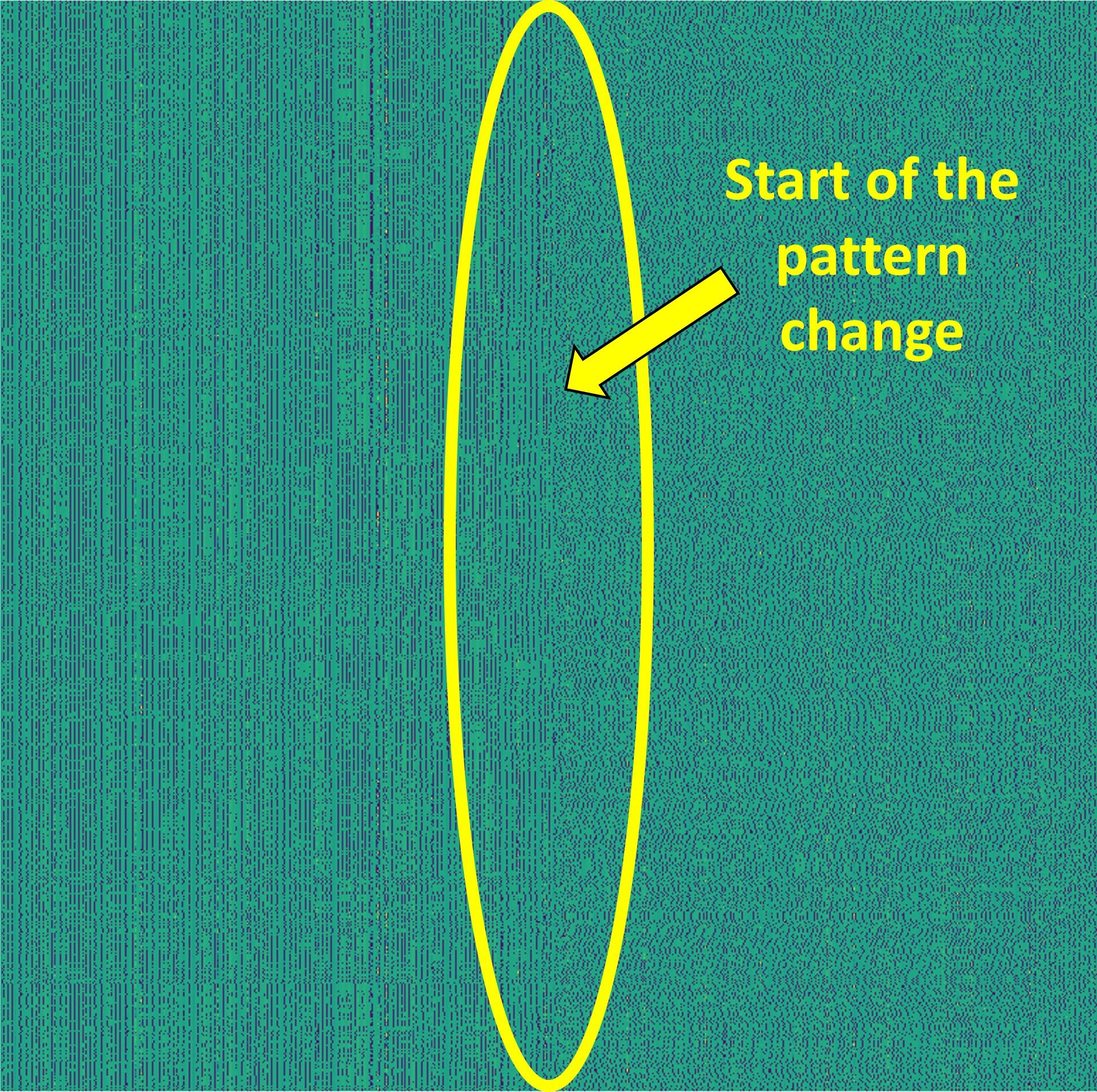}
    \caption{Temporal-spatial data matrices are converted into RGB images after preprocessing steps. The vertical axis represents the evaluation in time and the horizontal axis represents the position along the optical fiber in the images}
    \label{fig:TDSM}
\end{figure}

\subsubsection{Data Augmentation}
\label{sec:augmentation}
\begin{figure}[h]
    \centering
    \includegraphics[scale=1.5]{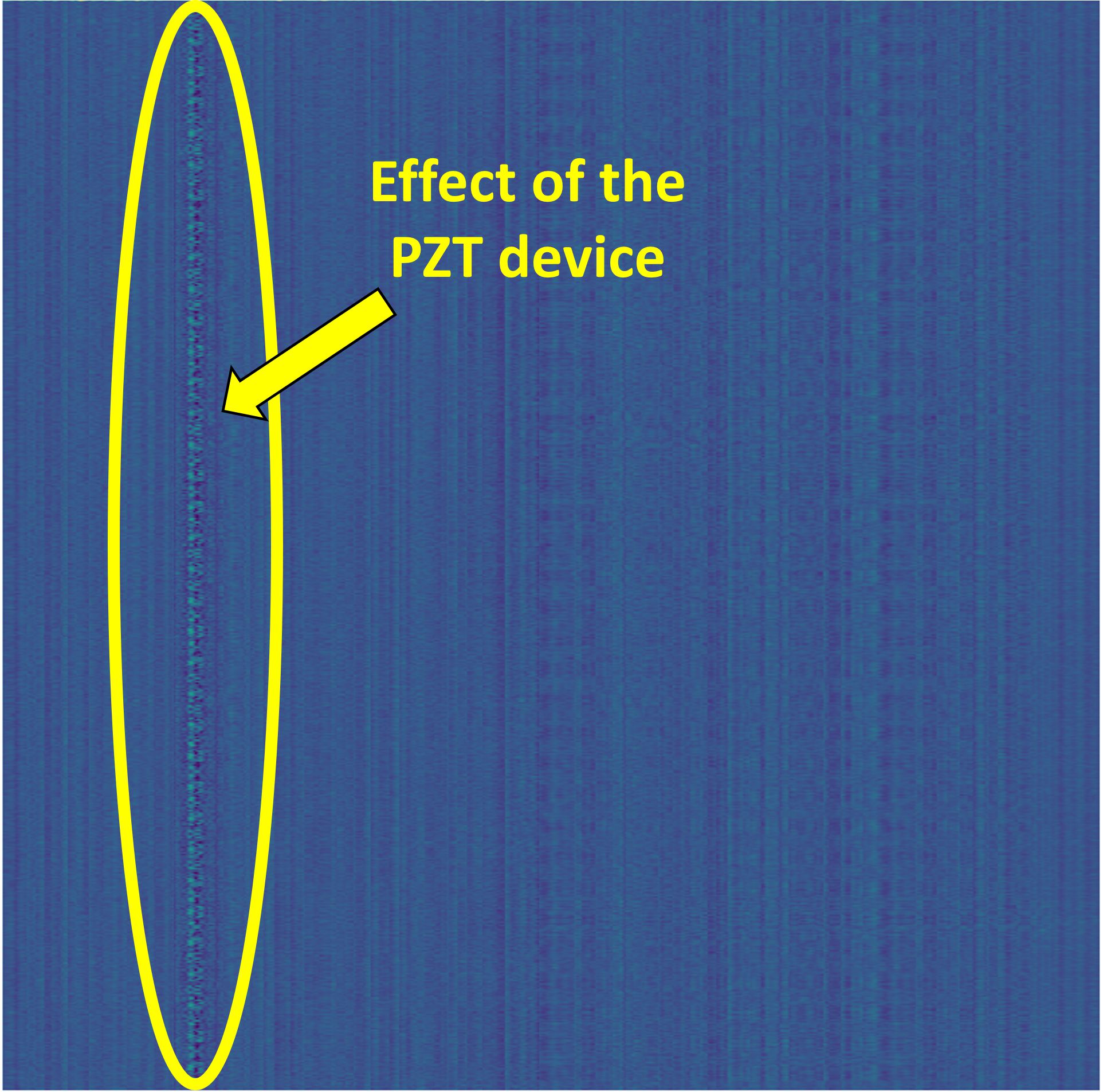}
    \includegraphics[scale=1.5]{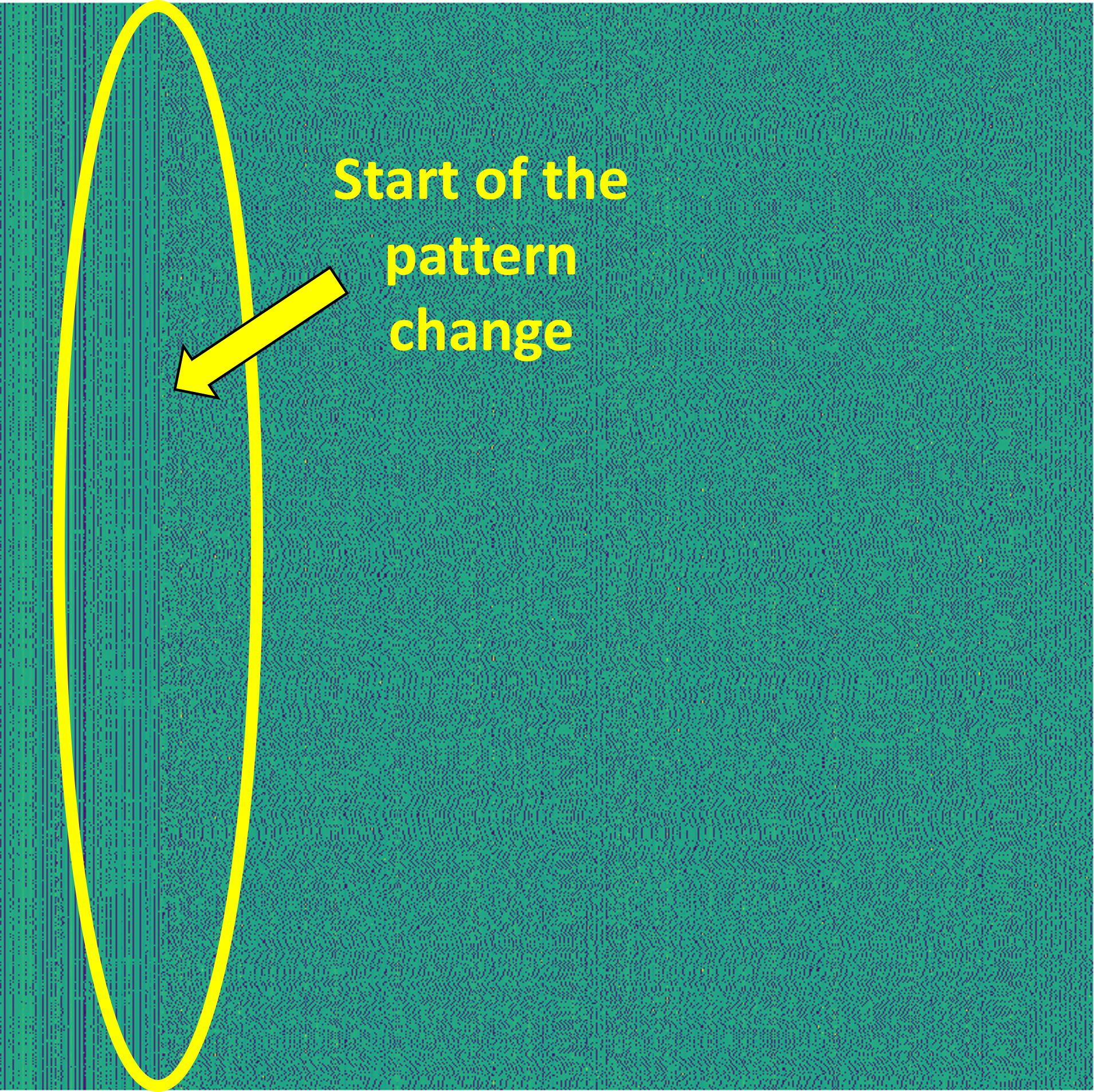}
    \caption{Temporal-Spatial data matrices obtained after the data augmentation steps}
    \label{fig:TDSM-AUG}
\end{figure}

To increase the size of the dataset, two different types of data augmentations are applied. In the first augmentation, the piezoelectric transducer (PZT) location is changed, and the matrices are modified accordingly. In the $\Phi$-OTDR recordings, the effect of the PZT can be observed in the region between the \nth{2320} and the \nth{2360} meters of the optical fiber. Therefore, the effect of the piezoelectric device can be observed in the \nth{23200} and the \nth{23600} columns in the temporal-spatial data matrix. To realize the data augmentation, the columns between the \nth{22000} and the \nth{25000} columns are replaced with the different 3000 adjacent columns in the data matrix. This operation has been applied nine times with different locations, and the number of image pairs is incremented from 15 to 150. The second data augmentation was to apply vertical flipping. The symmetry of each image around the x-axis is obtained and added to the dataset. By this methodology, the number of amplitude-phase image pairs is doubled and becomes 300. In the proposed framework, we used each amplitude-phase image pair as one sample and fed each sample to a different CNN. Differential amplitude and phase images with and without data augmentation are given in Figure \ref{fig:TDSM} and Figure \ref{fig:TDSM-AUG}.

\subsubsection{Computational Setup}
\label{sec:pcsetup}
The Python programming language was used for the preprocessing and training operations. We implemented our model in the Keras backend of the TensorFlow 2.6.0 framework. Computational experiments were performed on a laptop with the following properties: Intel i7-8750H CPU, 16 GB RAM, NVIDIA GeForce GTX 1060 4 GB GPU. We used the categorical cross-entropy loss function and the ADAM optimizer with its default settings. The batch size and number of epochs are selected as 16 and 25 respectively.

\subsection{Architecture of the Implemented Deep Learning Model}
\label{sec:modelarc}
\subsubsection{CNN}
\label{sec:cnn}
Convolutional Neural Network (CNN) is a widely used type of deep artificial neural network which has the capability of learning from data as in the image form. CNNs consist of feature extraction and classification sections. The feature extraction part of the network consists of stacked convolutional and pooling layers along with the classification part built by fully connected layers. Convolutional layers extract feature maps that are lessened by pooling layers after applying maximum or averaging operations \cite{10.5555/303568.303704}.
\paragraph{\emph{Convolutional Layer:}}

\label{sec:convlayer}
The convolutional layer is the fundamental layer of the CNN architectures. It extracts features from the high dimensional data by applying convolution operations with square matrices called kernels or filters.  Each convolutional layer outputs a feature map which is a three-dimensional tensor that stores the results of convolutions. The behavior of the ${n^{th}}$ convolutional layer in a CNN can be expressed by the following equation:

\begin{equation}
    \centering
    \large
    x_i=f\left(\sum_{j=1}^{N}W_k^{n-1}*y_m^{n-1}+b_k^n\right)  
\end{equation}

Where ${x_i^n}$ denotes the ${i^{th}}$ feature map, ${W_k^{(n-1)}}$ and ${b_k^n}$ represents the ${k^{th}}$ filters and the bias term, ${y_m^{(n-1)}}$ indicates the ${m^{th}}$ feature map in the previous layer, N is the number of total features and (*) denotes the convolution operation \cite{Albawi2017}. 

\paragraph{\emph{Pooling Layer:}}

\label{sec:pooling}

The pooling layer is applied to a feature map to be able to extract significant information by sliding a window over the feature map while applying the user-defined operation. It effectively decreases the size of the feature maps and passes only the most significant information. In this study, we used the max-pooling operation where only the maximum value in the sliding window is selected \cite{Srivastava2014}.

\paragraph{\emph{Fully Connected Layer:}}
\label{sec:fclayer}

The last section of a CNN is a feedforward neural network that is stacked using fully connected layers consisting of activation units. This section of CNN handles the classification of the extracted features in the previous layers. Activation units in the layers use their weights and the user-defined activation function to create an output. In this study, rectified linear unit (ReLU) activation function is used in each fully connected layer except the last layer. In the last layer, the softmax activation function is used to predict the output class of the images \cite{Lecun2015}. Mathematical descriptions of these two activation functions are given as follows:

\begin{equation}
\centering
\large
ReLU(x) = \left\{
        \begin{array}{ll}
            0 & \quad x \leq 0 \\
            x & \quad x \geq 0
        \end{array}
    \right. 
\end{equation}

\begin{equation}
    \centering
    \large
    Softmax(x_{i}) = \frac{e^{x_{i}}}{\sum_{k}{e^{x_{k}}}}
\end{equation}

\subsubsection{Transfer Learning and Pre-trained CNNs}
\label{sec:TL}
Training of a deep convolutional neural network is a process that requires a significant amount of resources such as large datasets and high-performance computation devices to overcome major problems such as avoiding overfitting and calculating partial derivatives propagating backwards. Additionally, they require user-supplied hyperparameter tuning for obtaining sufficient performance. Transfer learning is a methodology that is based on using pre-trained networks for avoiding these difficulties. In transfer learning, it is proposed to transfer a pre-trained network and replace some parts of it with new layers to train for fine-tuning. Pre-trained networks decrease the required time and resources for the training since they are already trained with intense resources and efforts. \cite{Lu2015,Pan2010}

\begin{figure}[h]
    \centering
    \includegraphics[scale=0.8]{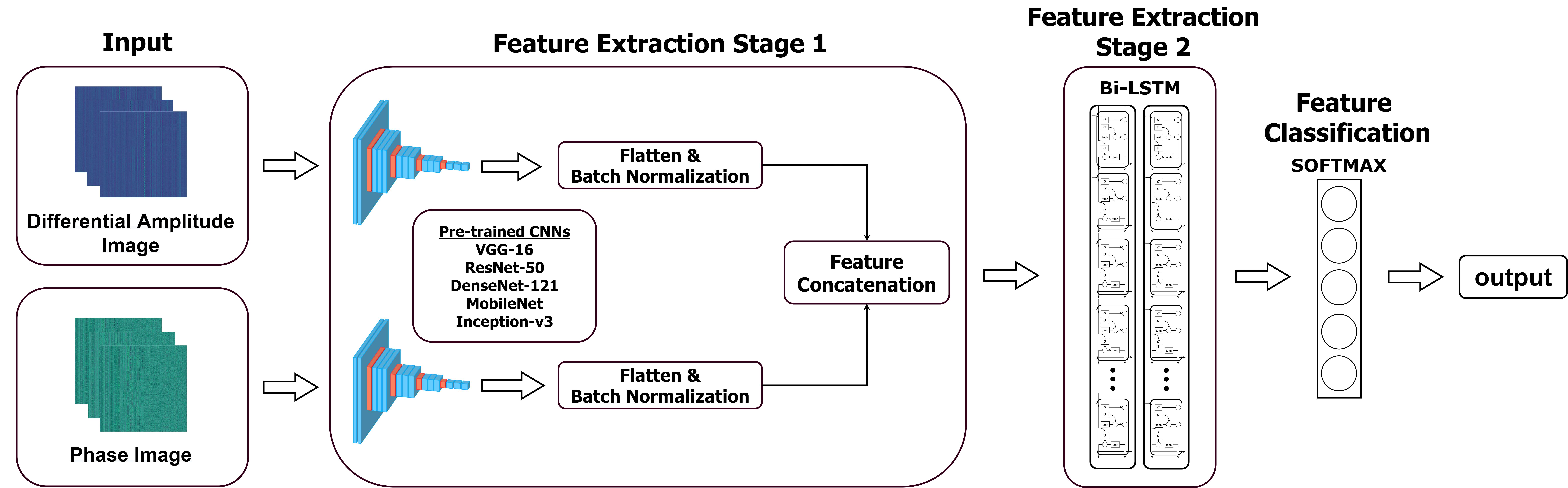}
    \caption{Schematic representation of the proposed framework for the classification of applied disturbances}
    \label{fig:algo-sketch}
\end{figure}

Pre-trained CNNs are networks that are previously trained on massive datasets by deep learning specialists. In this study, we constructed a CNN-based multi-input single-output model by using pre-trained CNNs. We utilized ResNet50, VGG16, DenseNet121, MobileNet, and Inception-v3 models as feature extractors during the study. These mentioned CNN architectures are examples of state-of-the-art pre-trained CNNs that are trained on the ImageNet dataset. ImageNet is a well-known dataset that consists of 14 million images from 20,000 different classes \cite{Russakovsky2015}. After feature extraction, we used long short-term memory (LSTM) layers to process the concatenated features which are extracted from the differential amplitude and phase images. The schematic representation of the proposed model is given in Figure \ref{fig:algo-sketch}.

\paragraph{\emph{ResNet-50:}}

\label{sec:resnet}
Residual Neural Network (ResNet) architecture is an improved CNN that is distinguished from a regular CNN by its shortcuts between layers. These shortcuts reduce the information distortion which is caused by the depth and complexity of the network. Additionally, it uses bottleneck blocks to reduce the amount of computational work to make training faster. ResNet-50 model is the 50-layer version of the ResNet architecture, and it has more than 23 million parameters. It is trained on the ImageNet dataset for image classification tasks \cite{He2015}.

\paragraph{\emph{VGG-16:}}

\label{sec:vgg16}
VGG16 architecture is an improvement over the famous AlexNet architecture which is the winner of the ImageNet Large Scale Visual Recognition Challenge (ILSVRC) in 2012. VGG16 architecture consists of five CNN-based feature extraction sections which are followed by stacked fully connected layers. Each feature extraction section ends with a max-pooling layer. All of the fully connected layers except the last one are equipped the ReLU activation function. In the last layer, the softmax activation function is used for classifying the extracted features. The total number of parameters in the VGG16 architecture is around 138 million. The VGG16 architecture is also trained on the ImageNet database \cite{Simonyan2014}.

\paragraph{\emph{DenseNet-121:}}
\label{sec:densenet}
DenseNet-121 is a state-of-art CNN architecture that consists of dense blocks which are modules that connect each layer with each other. Inside dense blocks, each of the previous feature maps are concatenated into a single tensor and used as inputs. The size of each feature map kept same inside a dense block for the concatenation. The downsampling operation is performed by transition layers. The transition layers connect the dense blocks by downsampling the output of the preceding dense block using convolution and pooling operations. DenseNet-121 architecture consists of 120 convolutional layers and about 7.6 million number of parameters. 

\paragraph{\emph{MobileNet:}}

\label{sec:mobilenet}
MobileNet is a well-known and efficient CNN architecture that is used in image classification applications. In MobileNet architecture, standard convolutions are replaced with depthwise separable convolutions to reduce the size of the model. One of the MobileNet’s distinguished features is that the MobileNet’s characteristic new hyperparameters (width and resolution multipliers) allows researchers to trade off latency for high training speed and less disk space. The MobileNet architecture consists of 13 million number of parameters \cite{howard2017mobilenets}.

\paragraph{\emph{Inception-v3:}}

\label{sec:inception}
Inception-v3 is a CNN architecture that consists numerous amounts of stacked and repeating inception modules which are followed by fully connected layers. These inception modules are parallel convolutional layers which lead to a concatenation layer. This architecture reduces the number of connections in the network without degrading the efficiency and performance of the model. In the end, it outputs a tensor with size 8x8x2048 and uses average pooling, dropout, and fully connected layers to make classification. The Inception-v3 architecture contains approximately 24 million number of parameters \cite{Szegedy2015}.

\subsubsection{LSTM}
\label{sec:LSTM}
Recurrent Neural Network (RNN) is a type of neural network which is capable of learning from sequential data. Long-short term memory (LSTM) network is a special kind of RNN which is generally used in sequential data applications such as, denoising, machine translation, speech recognition etc. It consists of stacked blocks that process the information transferred from the previous block with the present input to create an output. There are different gate structures in the LSTM blocks to update the output properly. In order to create suitable values, LSTM blocks use two type of activation functions which are sigmoid and tanh functions. Sigmoid function creates outputs within the range between zero and one which define the amount of information that is updated or forgotten. The LSTM architecture is designed to solve the infamous vanishing gradient problem which is the disappearing of the partial derivatives during the backpropagation. Since the LSTM architecture manages to overcome this problem, it is a much better methodology for capturing the long-term dependencies in the sequential data \cite{Hochreiter1997}. 

Bidirectional LSTM is an enhanced version of the LSTM architecture. The of the bidirectional LSTM layers are capable of analyzing the sequential data in both directions of time. By analyzing the sequential data in both ways, bidirectional LSTM layers capture the long-term relations more efficiently compared to regular LSTM layers. However, the analysis quality that comes from the bidirectionality has a certain trade-off. Bidirectional LSTM layers contains twice as many parameters compared to a regular LSTM layer and require more computational demand \cite{Schuster1997}. 

\begin{figure}[h]
    \centering
    \includegraphics[scale=0.4]{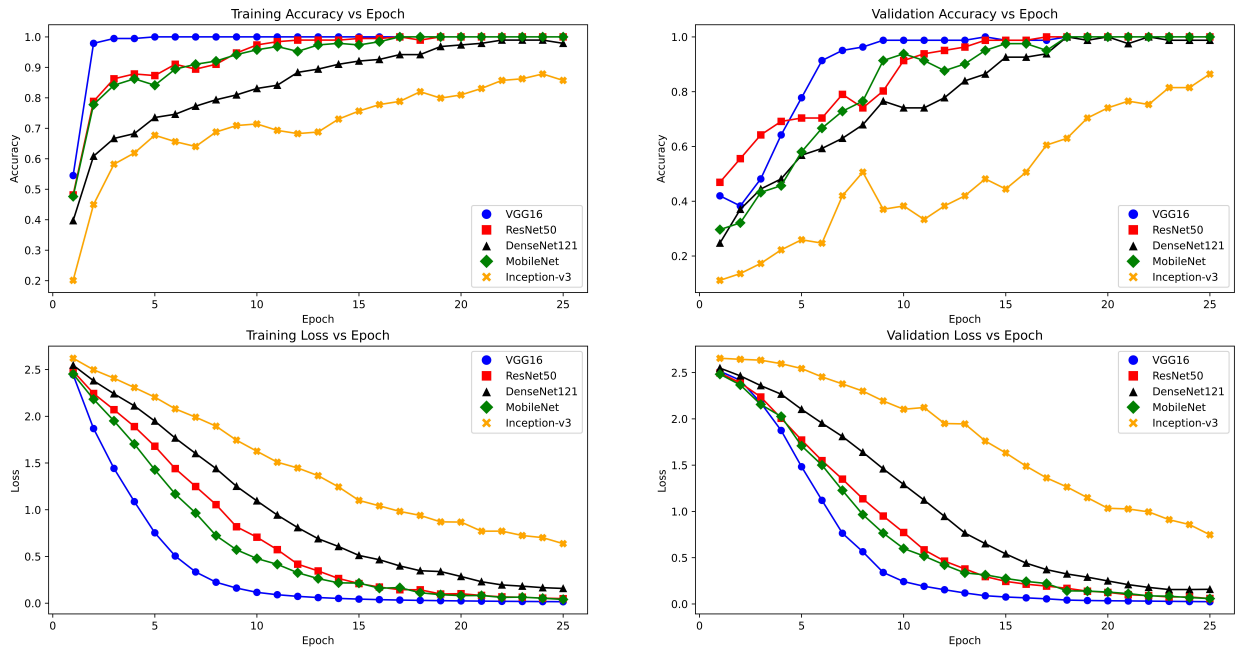}
    \caption{Evaluation of loss and accuracy metrics for randomly selected training from 50 runs}
    \label{fig:loss_acc}
\end{figure}

\section{Results}\label{sec2}

In this research, we built a multi-input single-output model by employing transfer learning and bidirectional-LSTM to classify the sinusoidal disturbances applied to an optical fiber. We achieved an accuracy value of 100\% by using pre-trained CNN architectures as feature extractors. The size of the dataset utilized in the study was insufficient to train deep CNNs efficiently. However, we utilized the concept of transfer learning to compensate for this limitation and managed to avoid the common difficulties of training CNNs such as the requirement of large datasets and enormous computational power. We removed the classifier stages of five pre-trained CNN architectures and used each CNN as a feature extractor. By this implementation, we kept the features extracted by CNN untouched and prevented any information loss. The architectures we employed are ResNet50, VGG16, DenseNet-121, MobileNet, and Inception-v3.

We completed the feature extraction in two stages since the data matrix we constructed contains both temporal and spatial information. In the first stage, we used a pre-trained CNN to extract the spatial features from each image. While loading the images, we did not apply any normalization. So, in order to improve the performance of our model, we applied batch normalization to the spatial features extracted by the CNNs. Then, the normalized feature vectors are merged to obtain a single normalized spatial feature vector. In the second stage of the feature extraction, we used two bidirectional LSTM (Bi-LSTM) layers to extract time-related information from the constructed feature vector since LSTM layers are efficient tools for analyzing sequential data without facing the vanishing gradients problem. We selected to use Bi-LSTM layers instead of regular LSTM layers since they are more capable of analyzing sequential data and can process the input sequence in both directions. Performance curves obtained after a ra<ndomly selected training are given in Figure \ref{fig:loss_acc}.

\begin{figure}[H]
    \centering
    \includegraphics[scale=0.7]{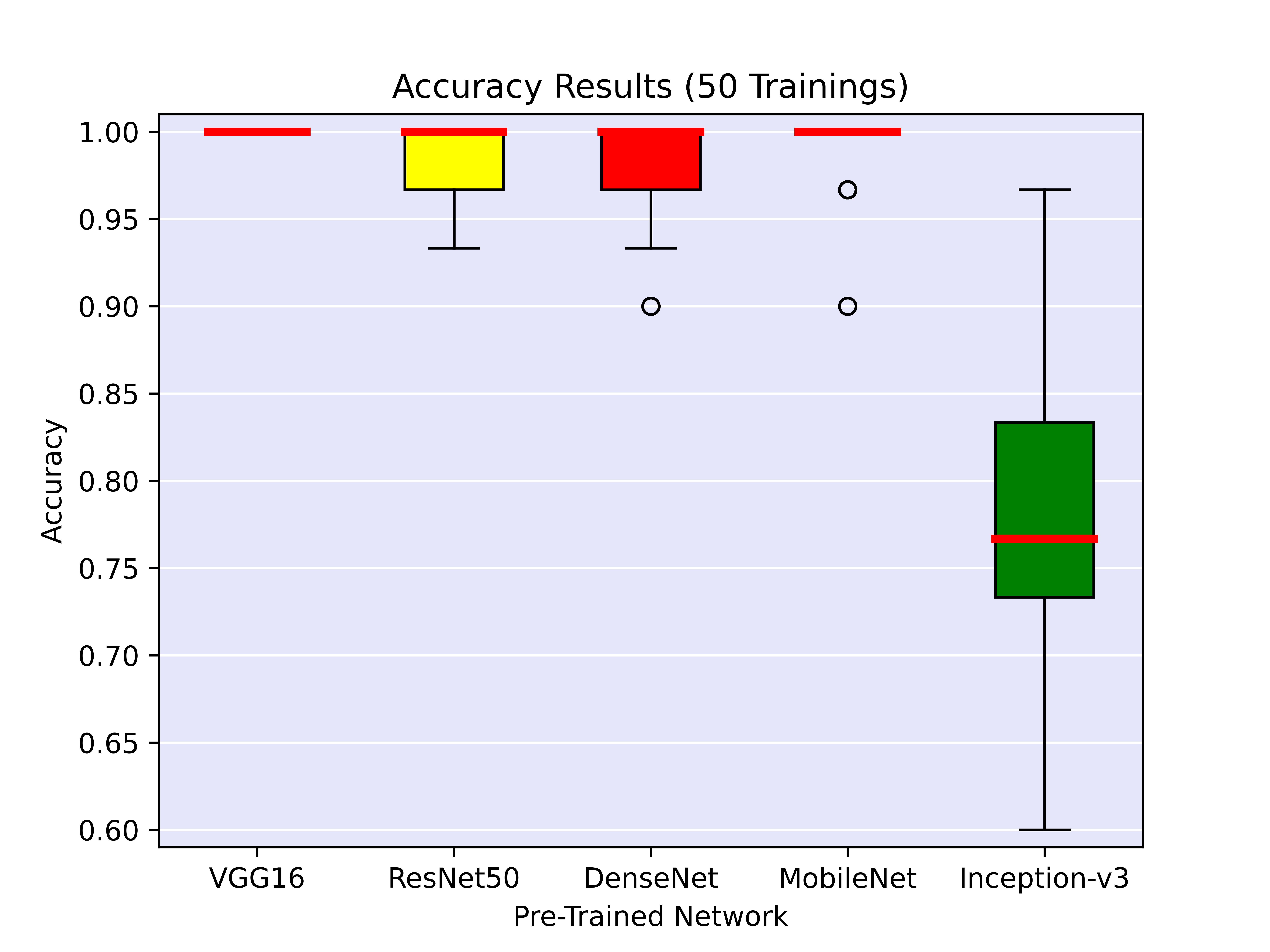}
    \caption{Boxplot representation of classification accuracy values on the test set for 50 trainings}
    \label{fig:box}
\end{figure}

During the experimental stage, 15 ${\Phi}$-OTDR recordings were recorded. With the help of the data augmentations mentioned in Section 2, we increased the number of samples for each class from one to twenty. So totally, we worked with a dataset of 300 data samples. We randomly separated the data in each run and trained our model with 70\% of our dataset which corresponds to 210 images.The boxplot that represents the accuracy distribution of 50 trainings for different feature extractors is given in Figure \ref{fig:box}. Some of the example confusion matrices are given in Figure \ref{fig:cm}. 

The results represented in Figure \ref{fig:box} shows that our methodology manages to obtain significant classification accuracy on the data. The mean accuracy values indicate that the setup that uses VGG16 architecture as a feature extractor performs significantly better. The differences in maximum and minimum accuracy indicate that the VGG16 architecture is also more consistent compared to other CNN architectures. In order to monitor the effect of our feature extraction stages, we constructed feature embeddings for the output of each feature extraction stage. This methodology allowed us to observe the discriminative quality of our features in the 2D space. Further analysis of the feature embeddings is explained in detail in the Discussions section. In summary, we managed to classify 15 different sinusoidal signals with different amplitude and frequencies using the proposed transfer learning-based framework.

\section{Discussion}\label{sec12}

The distributed acoustic sensors are widely used advanced devices that can make static and dynamic measurements along very long distances. Analysis of data collected from those kinds of applications requires advanced signal processing techniques with high computational demands. This requirement creates a serious issue for real-time analysis that contain outcome representation and classification. Recently, the artificial intelligence-based approaches, in particular, the deep learning methodologies improved the solutions for the related problems such as interpretation and classification \cite{Liehr2021}. 

The utilization of deep learning-based systems for optical data analysis is a topic with increasing popularity. Significant research on this subject is conducted in the literature. Sun $\emph{et al.}$ used a temporal-spatial data matrix to extract morphological features. Using the RVM classifier, they reached an average accuracy of 97.8\% \cite{Sun2015}. Shi $\emph{et al.}$ converted temporal-spatial data matrices into grey-scale images and used a CNN to classify 5 different events. They fine-tuned the famous Inception-v3 architecture and reached 96.67\% accuracy \cite{Shi2019}. Wu $\emph{et al.}$ processed the phase information alongside intensity and showed that the utilization of phase information increases the classification accuracy. They reached more than 85\% average classification accuracy \cite{Wu2021}. Liehr $\emph{et al.}$ utilized CNNs to develop a real-time denoiser and managed to outperform the BM3D denoiser \cite{Borchardt2020}.

\begin{figure}[H]
    \centering
    \includegraphics[scale=1.5]{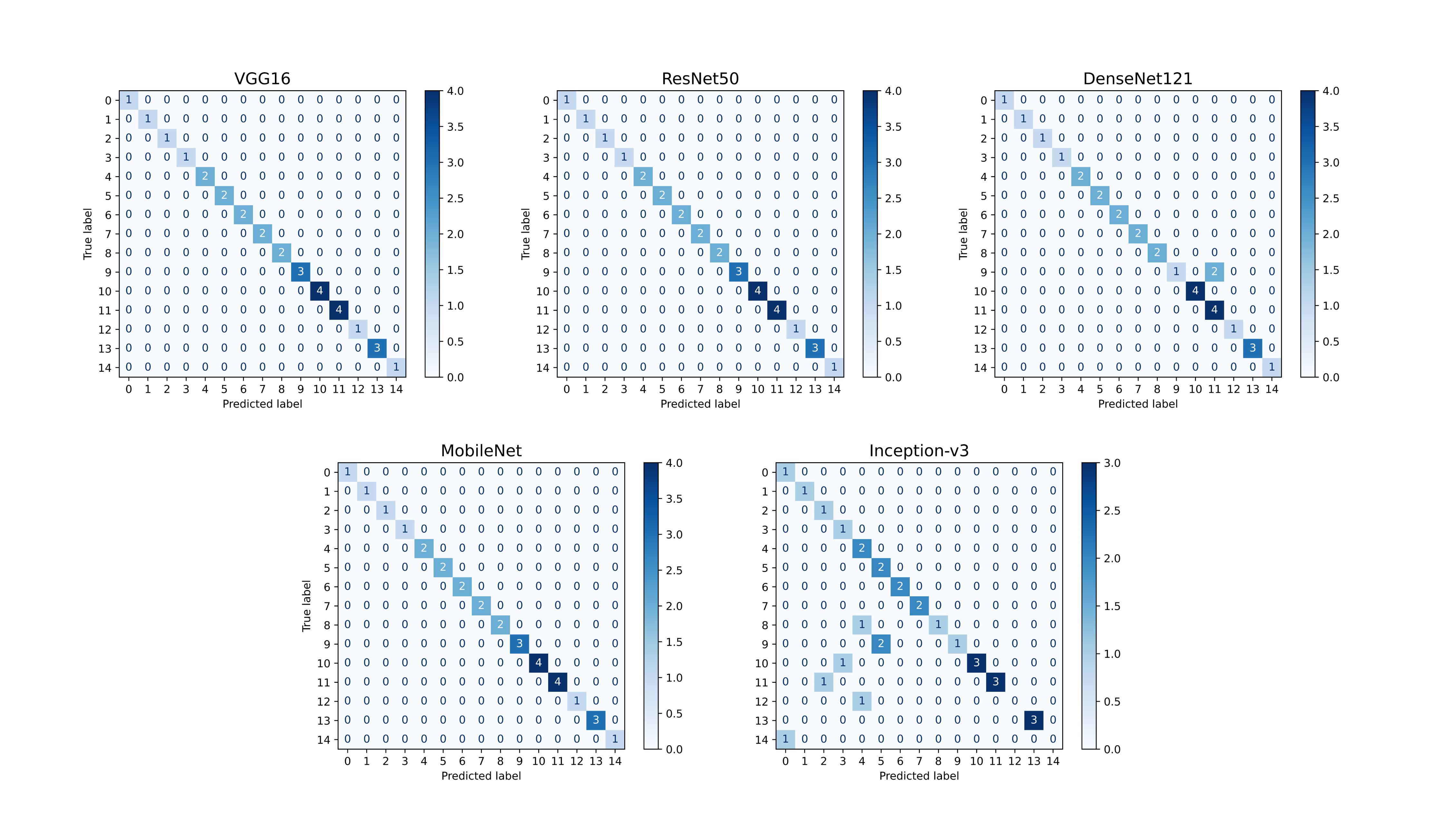}
\caption{Confusion matrices obtained after a randomly selected training from 50 runs}
    \label{fig:cm}
\end{figure}

Using pre-trained neural networks has been frequently proposed recently. With the help of the transfer learning tool, researchers managed to develop models that extract sufficient features with limited data. Narin $\emph{et al.}$ employed a pre-trained CNN for extracting features from chest X-Ray images and classified the extracted features with a feedforward artificial neural network. Using this methodology, they reached an average accuracy of 99.5\% on the test set with limited data \cite{Narin2021}. Shalbaf $\emph{et al.}$ used a pre-trained CNN followed by an SVM classifier for the detection of schizophrenia \cite{Shalbaf2020}. The detection and classification of events was the main task of the mentioned studies. However, the number of studies in the optical data analysis literature that is based on transfer learning is limited. Recently, Li $\emph{et al.}$ used the pre-trained AlexNet architecture for the classification of 8 events \cite{Li2022,Shi2019}.

\begin{figure}[h]
    \centering
    \includegraphics[scale=0.7]{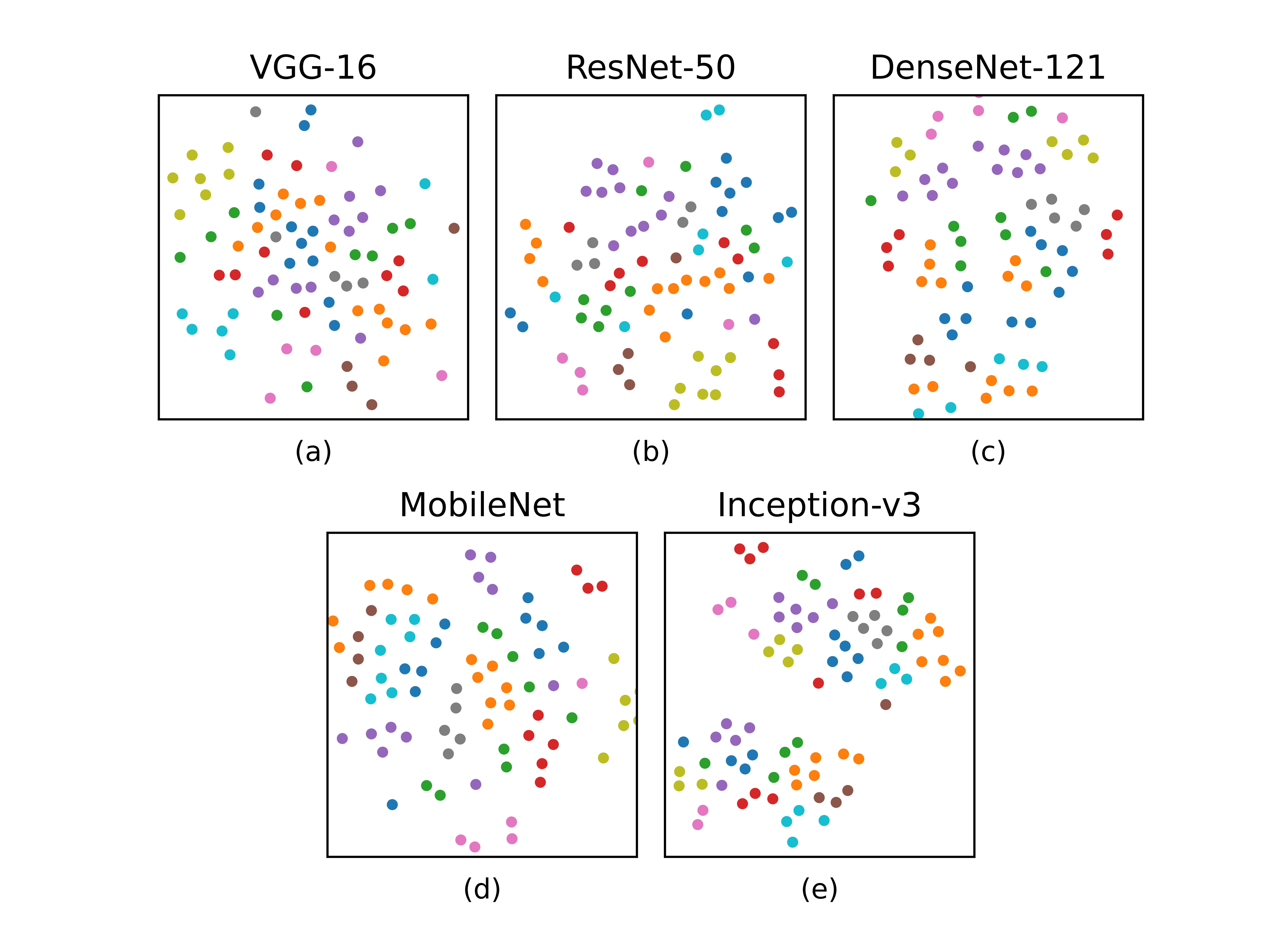}
    \caption{Feature embeddings of the extracted features after Feature Extraction Stage-1 (a) VGG-16, (b) ResNet-50, (c) DenseNet-121, (d) MobileNet, (e) Inception-v3}
    \label{fig:FES}
\end{figure}

\begin{figure}[h]
    \centering
    \includegraphics[scale=0.7]{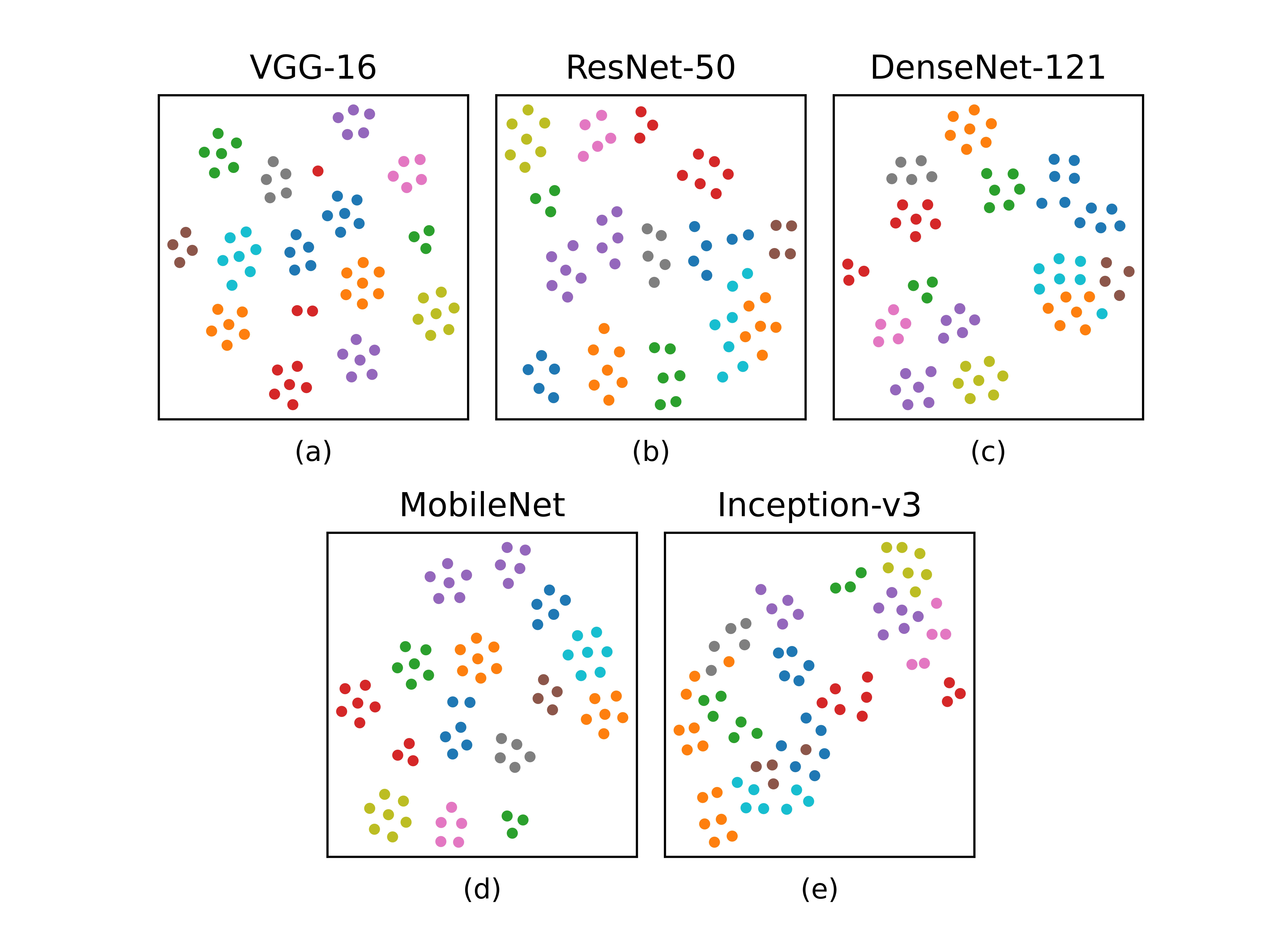}
    \caption{Feature embeddings of extracted features after Feature Extraction Stage-2 (a) VGG-16, (b) ResNet-50, (c) DenseNet-121, (d) MobileNet, (e) Inception-v3}
    \label{fig:FET}
\end{figure}

The main goal of this study is to show that the pre-trained CNNs can extract highly effective features from the optical data represented in a 2D form. Additionally, we wanted to demonstrate that different types of information can be learned from the data using different type of feature extractors if they are used effectively and properly. Our study differs from previous works in terms of methodology since we are using both phase and amplitude information which are represented in the form of temporal-spatial data matrices. Additionally, we removed the classifier modules of the pre-trained CNNs and applied batch normalization to the extracted features. Moreover, we employed Bi-LSTM layers for further extraction of temporal information. As Figure \ref{fig:FES} and Figure \ref{fig:FET} show, our 2-stage feature extraction methodology allowed us to obtain very high-quality features which is a novel methodology in the optical domain research.

The most important points in the study can be expressed as follows:
\begin{itemize}
    \item Feature extraction, selection, and classification processes are completed automatically.
    \item 15 sinusoidal disturbances applied on optical fiber are classified.
    \item Feature extraction is completed with pre-trained CNNs and the performances of 5 different state-of-the-art CNN architectures are evaluated and compared. 
    \item Bi-LSTM layers are utilized for extracting temporal information from the spatial features.
    \item High classification accuracy with a limited amount of data is obtained.
    \item A new framework based on 2-stage feature extraction is proposed.
\end{itemize}

There is a data availability problem in the optical domain. The transfer learning methodology enables to overcome the inadequacy of using a small dataset which limits the training of enormous number of parameters of deep CNN architectures. It’s not possible to compare our framework directly with other studies since there is unfortunately no publicly available benchmarking dataset in this research area. In order to contribute to the literature, we are sharing our dataset to observe the performance of studies with different approaches. 

As Figures \ref{fig:box}-\ref{fig:FET} show, the classification accuracy of our framework strongly depends on the CNN architecture that is employed for the extraction of spatial features. Internal mechanisms of the employed CNN architectures are briefly mentioned in the Methods section. Considering the given information about the architectures with the accuracy results, we have reached the following outcomes:
\begin{itemize}
    \item The VGG-16 architecture managed to extract better features compared to other architectures because of its 13 convolutional layers and the enormous number of parameters. When the created temporal-spatial data matrices are checked, some oscillations along the vertical axis can be observed for both images. These oscillations are capable of identifying each of our sinusoidal disturbances. We believe that the convolution-based direct approach of the VGG-16 architecture manages to identify these patterns of different classes in our temporal-spatial data matrices.
    \item The ResNet-50 architecture uses residual connections which results in the processing of a feature map with the original data. We expressed the imperfect classification accuracy of ResNet-50 as the information loss caused by the possible domination of the original information over the extracted features.
    \item Inside its dense blocks, the DenseNet-121 architecture contains all of the feature maps created previously, which are processed by the convolutional layers. Unlike ResNet-50, DenseNet-121 does not use the unprocessed data with the features. Instead, it uses dense connections that connect the output of each layer with the input of the next layers. We expressed the classification success of the DenseNet-121 by this homogenous processing of the information stored in the temporal-spatial data matrix. 
    \item Since the MobileNet architecture uses depthwise convolutions instead of standard ones, we believe that the high classification accuracy of MobileNet is caused by its discrimination capability of the patterns on the temporal-spatial data matrix with the constant background.
    \item The Inception-v3 architecture consists of the repetition of several inception modules which are connected by grid-size reductions. As our results indicate, the Inception-v3 architecture fails to classify our data samples. We believe that the inception modules which are the building blocks of the architecture are inappropriate for the extraction of sufficient information from the temporal-spatial data matrices which causes the extraction of insufficient features repetitively.
\end{itemize}

In order to represent the performance of our methodology explicitly, we created feature embeddings using t-distributed stochastic neighbor embedding (t-SNE) for the features learned after each feature extraction stage \cite{VanDerMaaten2008}. t-SNE is a dimensionality reduction technique that is mainly used for the visualization of high-dimensional data. It projects the data into a two-dimensional (2D) plane. Distance between two data points in the 2D plane is determined by their distance in the original space. From this fact, it can be stated that an unsatisfactory representation in the 2D space does not always mean that the data is inadequate, however, a well-represented visualization in the 2D space certainly indicates well-organized data. 

By comparing the feature embeddings of the first and second feature extraction stages given in Figure \ref{fig:FES} and Figure \ref{fig:FET}, advancement in the features after the Bi-LSTM layers can be easily observed. This macroscopical enhancement in the features supports our main proposal, which was the extraction of additional, temporal information by the Bi-LSTM layers. Considering the initial and final feature embeddings it can be clearly stated that the Bi-LSTM layers managed to extract the time-related information from the spatial features learned in the first stage.

In order to observe the effect of transfer learning comprehensively, we tested our framework with five state-of-the-art CNN architectures, which embrace different approaches to the image classification task. Alongside the VGG-16 architecture, which makes a more straightforward approach with its enormous number of parameters, we employed different models that utilize unique analysis techniques such as residual connections, dense blocks, depthwise convolutions, and repeating inception modules. Figure \ref{fig:FES} and Figure \ref{fig:FET} also represent that the feature embeddings also contain information about the compatibility of CNN architectures for this task. In Figure \ref{fig:FES}, it can be clearly seen that the spatial features are not enough for proper discrimination. However, in Figure \ref{fig:FET}, classes are represented very clearly except for the features of Inception-v3. Additionally, they are so distinct that even classical machine learning methods such as KNN, SVM, or K-means would be sufficient for classifying the samples accordingly. These transformations of the features also indicate that the spatial features extracted with pre-trained CNNs are also compatible with Bi-LSTM layers. However, the embeddings related to Inception-v3 clearly represent that extracted features are neither sufficient for classification nor suitable for Bi-LSTM layers.

\section{Conclusion}\label{sec13}

Artificial intelligence-based smart decision frameworks for distributed acoustic sensing systems are promising significant benefits for the optical domain. In this study, we proposed an event classification methodology for $\Phi$-OTDR distributed acoustic sensing system based on deep transfer learning. We extracted the amplitude and phase information from the data and constructed temporal-spatial data matrices for each information. In order to classify the temporal-spatial data matrices, we implemented a multi-input single-output deep learning model that extracts features in two stages, using pre-trained CNN architectures and bidirectional LSTM layers. In the first stage, we extracted the spatial information by using pre-trained CNN architectures. In the second stage, we stacked the spatial information extracted from amplitude and phase images and constructed a feature vector. In order to retrieve the time-related information from the extracted features, we used two Bi-LSTM layers. Our results show that the VGG-16 architecture yielded 100\% accuracy which is the highest among the five CNN architectures we employed. The feature embeddings obtained after each feature extraction stage show that the implemented Bi-LSTM layers perfected the extracted features. Thanks to transfer learning, we managed to train our framework with a relatively small dataset without high computational demand. Additionally, we showed that the state-of-the-art pre-trained CNN architectures are suitable for extracting sufficient information from images which are unrelated to the training dataset. Besides, we managed to avoid the coherent fading phenomenon. One significant observation from this study was the performance of our framework with different pre-trained CNN architectures. Our results implied that the VGG-16 architecture is more capable of extracting spatial information from the data with its enormous number of parameters. It is also observed that the characteristic problem-solving techniques of other CNN architectures are limiters for extracting features that are proper for LSTMs.

\section{Acknowledgments}

Kivilcim Yüksel acknowledges financial support from the Scientific and Technological Research Council of Turkey, (TUBITAK, BIDEB-2219-1059B191600612). $\Phi$-OTDR measurements were taken  at the University of Mons, Electromagnetism and Telecommunications Department, under the supervision of Prof. Dr. Marc Wuilpart.

\bibliographystyle{unsrt}  
\bibliography{references}  

\end{document}